\documentclass[conference]{IEEEtran}
\IEEEoverridecommandlockouts
\usepackage{cite}
\usepackage{amsmath,amssymb,amsfonts}
\usepackage{algorithmic}
\usepackage{graphicx}
\usepackage{textcomp}
\usepackage{xcolor}
\usepackage{siunitx}
\usepackage{soul}
\usepackage{booktabs}

\usepackage{siunitx}
\usepackage{soul}

\makeatletter
\newcommand{\linebreakand}{%
  \end{@IEEEauthorhalign}
  \hfill\mbox{}\par
  \mbox{}\hfill\begin{@IEEEauthorhalign}
}
\makeatother

\def\BibTeX{{\rm B\kern-.05em{\sc i\kern-.025em b}\kern-.08em
    T\kern-.1667em\lower.7ex\hbox{E}\kern-.125emX}}
\begin{document}

\title{
    Hardware-Algorithm Re-engineering of Retinal Circuit for Intelligent Object Motion Segmentation
}

\author{\IEEEauthorblockN{Jason Sinaga*\thanks{*These authors contributed equally to this work.}}
\IEEEauthorblockA{\textit{Electrical and Computer Engineering} \\
\textit{University of Wisconsin---Madison}\\
Madison, USA \\
sinaga@wisc.edu}
\and
\IEEEauthorblockN{Victoria Clerico*}
\IEEEauthorblockA{\textit{Electrical and Computer Engineering} \\
\textit{George Mason University}\\
Fairfax, USA \\
mclerico@gmu.edu}
\and
\IEEEauthorblockN{Md Abdullah-Al Kaiser}
\IEEEauthorblockA{\textit{Electrical and Computer Engineering} \\
\textit{University of Wisconsin---Madison}\\
Madison, USA \\
mkaiser8@wisc.edu}
\linebreakand
\IEEEauthorblockN{Shay Snyder}
\IEEEauthorblockA{\textit{Electrical and Computer Engineering} \\
\textit{George Mason University}\\
Fairfax, USA \\
ssnyde9@gmu.edu}
\and
\IEEEauthorblockN{Arya Lohia}
\IEEEauthorblockA{\textit{Ideaventions Academy}\\
Reston, USA \\
alohia@ideaventionsacademy.org}
\and
\IEEEauthorblockN{Gregory Schwartz}
\IEEEauthorblockA{\textit{Feinburg School of Medicine} \\
\textit{Northwestern University}\\
Evanston, USA \\
greg.schwartz@northwestern.edu}
\linebreakand
\IEEEauthorblockN{Maryam Parsa}
\IEEEauthorblockA{\textit{Electrical and Computer Engineering} \\
\textit{George Mason University}\\
Fairfax, USA \\
mparsa@gmu.edu}
\and
\IEEEauthorblockN{Akhilesh Jaiswal}
\IEEEauthorblockA{\textit{Electrical and Computer Engineering} \\
\textit{University of Wisconsin---Madison}\\
Madison, USA \\
akhilesh.jaiswal@wisc.edu}
}

\maketitle

\begin{abstract}
Recent advances in retinal neuroscience have fueled various hardware and algorithmic efforts to develop retina-inspired solutions for computer vision tasks. In this work, we focus on a fundamental visual feature within the mammalian retina, Object Motion Sensitivity (OMS). Using DVS data from EV-IMO dataset, we analyze the performance of an algorithmic implementation of OMS circuitry for motion segmentation in presence of ego-motion. This holistic analysis considers the underlying constraints arising from the hardware circuit implementation. We present novel CMOS circuits that implement OMS functionality inside image sensors, while providing run-time re-configurability for key algorithmic parameters. In-sensor technologies for dynamical environment adaptation are crucial for ensuring high system performance. Finally, we verify the functionality and re-configurability of the proposed CMOS circuit designs through Cadence simulations in 180nm technology. 
In summary, the presented work lays foundation for hardware-algorithm re-engineering of known biological circuits to suit application needs. 


\end{abstract}


\section{Introduction}
The mammalian retina has evolved into a sophisticated visual processing system that can process parallel streams of visual information collected through photoreceptors and translate them into highly specific \textit{feature-spikes} \cite{Sernagor2001RGC}. Feature-spikes encode information in real-time about multiple visual features such as object motion, shape, or orientation. These visual features carry critical significance in animal survival, such as dictating stereotypical escape or attack maneuvers. Recent retinal neuroscience advances have led to a detailed understanding of specific biological circuits responsible for generating fundamental visual features \cite{schwartz2021object}. Advancements in retinal neuroscience have driven the development of hardware and algorithmic solutions trying to \textit{mimic} the processing and ability of retinal circuits.

Among neuromorphic sensors, Dynamic Vision Sensors (DVS) are the most well-known hardware effort to imitate mammalian retinal computations in novel camera systems and have revolutionized many visual sensing tasks such as optical flow estimation \cite{bartolozzi2014flow}, object tracking \cite{visevent}, and autonomous robot maneuvering \cite{vasco2017icub}. Traditional cameras, such as CMOS active pixel sensors and charge-coupled devices, rely on frame-based capture, where the entire scene is captured at a fixed rate, leading to excessive computation and bandwidth requirements \cite{CCD_cams}. In contrast, DVS cameras generate asynchronous spikes in response to changes in pixel intensity, leading to highly sparse data generated from the camera \cite{gallego2020event}. In moving-camera applications, such as autonomous maneuvering, DVS generates a burst of events due to platform shifts, and therefore, lacks direct applicability on ego-motion compensation tasks. 

Additionally, computations in DVS cameras are limited to photoreceptors and the first retinal synapse. Similarly, other methods worked on translating retinal computations into silicon technology included chips that emulated photoreceptor dynamics and the outer plexiform layer (OPL) \cite{siliconretina, siliconRetina2, siliconRetina3}. However, these implementations utilized phototransistors, which are not commonly used in commercial cameras. Their slow response further makes them unsuitable for high-speed applications (e.g., high-speed drones, car driving, etc.).   

Recently, Integrated Retinal Functionality in Image Sensors (IRIS) \cite{yin2022iris} has proposed a system aimed at emulating complete retinal computations, spanning from the input at the retina's photoreceptors to its output via retinal ganglion cells. In this work, we go beyond solely \textit{mimicking retinal circuits} to propose algorithm-driven hardware design \textit{re-engineering of retinal circuits} for object motion segmentation task in the presence of ego-motion. This paper focuses on the re-configurable retinal Object Motion Sensitive (OMS) circuit \cite{schwartz2021object}, with the following key contributions:

\begin{enumerate}
    \item Algorithmic Re-engineering: We present a holistic analysis of key parameters for the retinal OMS circuit - the ability of the retina to distinguish the self-motion of the head or body from the surrounding motion, recognizing the movements of objects. We show that optimizing (re-engineering) key OMS circuit parameters enables the added ability to segregate object motion based on the relative object size.
    \item Hardware Re-engineering: We present novel IRIS-compatible hardware circuits leveraging 3D hybrid integration of CMOS chips that enable real-time reconfiguration of the key OMS circuit parameters. This in turn enables OMS feature-extraction in camera systems while allowing adjustment of hardware circuit parameters guided by algorithmic results. The proposed 3D integrated hardware solution enables optimization of the hardware OMS circuit for intelligent object motion segmentation including the ability to segment object motion based on relative object sizes. 
    \item Finally, we accentuate the intricate bio-inspired hardware-algorithm co-design showcasing interconnections between accuracy, algorithmic optimization and its associated hardware overhead.
\end{enumerate}

\section{Related Work}
The retina is one of the core components of the human visual system, and is made of three main layers: the photoreceptor layer, the outer plexiform layer (OPL), and the inner plexiform layer (IPL) \cite{retina_layers}. Each layer plays a fundamental role in the computation of Object Motion Sensitivity (OMS), among over 40 additional visual features. The photoreceptor layer transduces visual stimuli into electrical signals. Then, bipolar cells in the OPL respond to luminance changes. For OMS computations, amacrine cells form connections between bipolar signals in the OPL and the Retinal Ganglion Cells (RGC) in the IPL. Through inhibitory and excitatory synapses, amacrine cells integrate contrast signals from a global area (\textit{surround}) and subtract them from the local area (\textit{center}). Consequently, RGCs corresponding to OMS circuit respond to motion in the local area and to differential motion between the global and local areas \cite{schwartz2021object}. Current neuromorphic vision sensors such as DVS and DAVIS are inspired by computations in the photoreceptor layer and OPL \cite{ref_DVS, ref_DAVIS}. In this work, we investigate hardware-algorithm re-engineering of the retinal OMS circuit comprising OPL and IPL functionalities. A brief introduction to related works on camera-compatible hardware and application-driven algorithmic implementation of the retinal OMS circuit is given below.

\begin{figure}[ht]
    \centerline{\includegraphics[width=80mm]{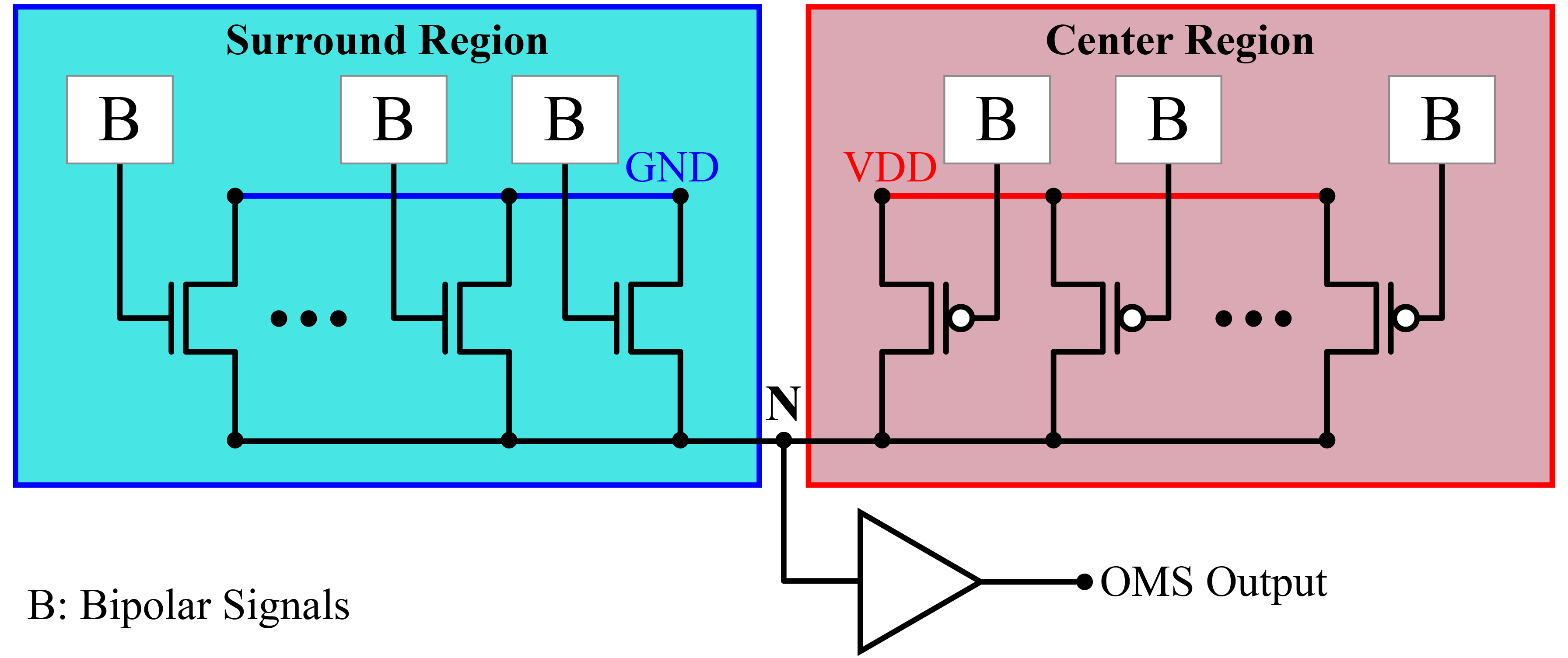}}
    \caption{Original IRIS circuit with set center and surround regions from \cite{yin2022iris}.}
    \label{fig:IRIS_circuit}
\end{figure}

Previous work proposed Integrated Retinal Functionality in Image Sensors (IRIS) \cite{yin2022iris}, a novel retina-inspired camera that implements full retinal computations throughout OPL and IPL, thereby executing retinal OMS computation seamlessly. In the initial CMOS design of OMS circuitry, NMOS (connected to GND) and PMOS transistors (connected to VDD) controlled by the bipolar signals were used for the surround and center regions respectively; see Figure \ref{fig:IRIS_circuit}. Depending on the motion of the visual scene, the surround and center region-specific transistors are modulating the overall charge on the node `N', see Figure \ref{fig:IRIS_circuit}.

Upon detection of abrupt motion within the center region, charge is injected into the node `N' through PMOS transistors. Conversely, the detection of such motion in the surrounding region triggers removal of charge from the node `N' through the NMOS transistors. Upon sufficient charge accumulation within the circuit, a positive voltage spike (\textit{OMS Output}) is generated, representing the detection of motion \cite{yin2022iris}.  A key limitation of the previous circuit is the inability to dynamically reconfigure the size and location of center and surround regions and other key parameters of the OMS circuit.

An algorithmic implementation of the OMS circuit derived from experimental neuroscience was presented in Snyder et al. \cite{OMS_CHANGE}. This method proposed a software algorithm based on convolutional kernels to compute OMS from DVS data and distinguish object motion from camera motion (ego-motion). This previous work provided a quantitative comparison with several state-of-the-art methods for ego-motion compensation and object motion segmentation. 

In this work, we combine prior works on hardware and algorithmic implementation of the OMS circuit to propose novel re-engineering of OMS circuit across hardware and algorithm.

\section{Methodology}
The Object Motion Sensitivity (OMS) algorithm proposed in \cite{OMS_CHANGE}, grounded in experimental neuroscience, offers a robust framework for object motion segmentation in the presence of ego-motion. The algorithmic implementation was designed to be tunable, allowing for application-specific configuration adjustments to optimize performance across diverse environments. While the original configuration provides improved performance for the used ego-motion datasets, it failed to account for the overhead associated with hardware circuit implementation. Consequently, we study the effects of various key parameters on the algorithm's performance and their relation to the hardware implementation. The software algorithmic analysis is then used to guide CMOS circuit design which enables an OMS circuit that is re-configurable during runtime.

\subsection{Software implementation}
Previous work proposed an algorithmic implementation of the OMS biological circuitry derived from experimental neuroscience \cite{OMS_CHANGE}. The OMS algorithm, tested on synthetic and real DVS data, aimed to functionally replicate neural computations performed by the amacrine and retinal ganglion cells (RGC).

The OMS algorithm from \cite{OMS_CHANGE} takes as input the photoreceptor activations of bipolar cells, represented by DVS frames. This algorithmic implementation consists of two convolutional squared filters; the \textbf{center kernel} and the \textbf{surround kernel} representing the center and surround regions from the human visual system. The largest of the kernels (surround) acts as the connection between the bipolar cells and the amacrine cells, while the smallest kernel (center) serves as the synapse between RGCs and their corresponding bipolar cell cluster.  

The kernels convolve over each frame by centering themselves over each pixel and storing the average resultant value in the position of the given pixel. To simulate the inhibitory synapses, the mean contrast values from the surround (amacrine) filter are subtracted from those of the center (RGC) filter. Afterward, if the resultant values are larger than a given threshold, $\tau$, a boolean spike is stored in the OMS output frame for said pixel. 
Table \ref{tab:params} summarizes the tunable parameters associated with a specific configuration to compute OMS and their default values from the original paper \cite{OMS_CHANGE}. 

\begin{table}[h]
\centering
\caption{Tunable parameters for the OMS algorithm~\cite{OMS_CHANGE}.}
\label{tab:params}
\begin{tabular}{cc}
\toprule
Parameter              & Value           \\
\midrule
kernel type            & Gaussian sphere \\
center kernel shape (\textit{cen})    & 4x4             \\
surround kernel shape (\textit{surr})  & 8x8             \\
surround kernel stride ($s$) & 1               \\
threshold ($\tau$)        & 0.96           \\
Time interval ($T_s$) & 20 ms \\
\bottomrule
\end{tabular}
\end{table}

The \textbf{center and surround kernels} correspond to two Gaussian filters with radius $\frac{cen}{2}$ and $\frac{surr}{2}$ respectively. The weights are sampled from a normalized Gaussian distribution. Due to their effectiveness in suppressing noise, Gaussian filters are frequently employed as a pre-processing step for smoothing images in convolutional neural networks \cite{gauss_kernel}. The \textbf{surround kernel stride} $s$ was set to 1 to maintain the input's original dimensions at the output. The ground truth labels for the object motion segmentation task consist of binary images that provide a per-pixel motion segmentation. Therefore, maintaining output sizes that match the ground truth masks enables direct comparison between the results and the ground truth. 

Additionally, the shapes of the center and surround kernels and the threshold value were chosen for the specific dataset and task. The Gaussian center kernel was designed with a radius of 2 (i.e. a 4x4 matrix) to avoid covering an entire object, whereas the Gaussian surround kernel radius was set to 4 (i.e. an 8x8 matrix) to cover a broader area and detect the spikes in the global area effectively. Similarly, the \textbf{threshold} was set to a relatively large value because of the high density of spikes from background movement. However, none of these parameters were assessed for different scenarios. 

As mentioned above, the inputs to the algorithm are DVS frames. However, DVS cameras provide a burst of binary events containing polarity, X-coordinate, Y-coordinate, and capture time \cite{gallego2020event}. Hence, two-dimensional frames were created for each sequence out of a burst of events. This pre-processing step involved compressing the polarity axis and accumulating the events within a \textbf{time interval ($T_s$)}. The events within a neighborhood of 10 milliseconds around the ground truth mask capture time were accumulated into a single frame, resulting in a time interval of 20 milliseconds. The time interval directly impacts the frames' spike density, whereas large time windows result in a higher spike density. Therefore, adjusting the time window requires adjusting the remaining algorithmic parameters to optimize performance.

Optimizing the above-mentioned parameters for different scenarios and constraints is studied in this paper. We aim to provide a parameter configuration that accommodates the corresponding hardware design while guiding parameter re-configuration in the hardware for   varying environment conditions and tasks.

\subsection{Reconfigurable CMOS OMS Circuit}

The algorithm outlined earlier introduces a series of key parameters in Table \ref{tab:params} that serve as the foundation for designing a reconfigurable circuit tailored for OMS. Furthermore, we aim to provide CMOS-based OMS circuits with added capability for run-time reconfigurability of the key OMS parameters, specifically \textbf{kernel size} and \textbf{threshold}. 
Due to inherent leakage in the CMOS circuits, the passive capacitors cannot retain accumulated charges for long time window. Though utilizing intricate circuit design techniques, moderately long (in ms range) integration time can be achieved \cite{leakage_cmos, leakage_nvm}, however, hardware favors short integration times. In addition, low stride values (high overlapping) and complex kernel type require multiple OMS compute transistors per DVS pixel, which potentially increases the routing complexity and can add overhead in the pixel pitch (pixel density) requirement. These hardware constraints dictate the optimization of the algorithmic parameters (\textbf{stride, kernel type, threshold and time integration window $T_s$}), leading to an intricate hardware-algorithm co-design solution.

\begin{figure}[ht]
\centerline{\includegraphics[width=0.95\linewidth]{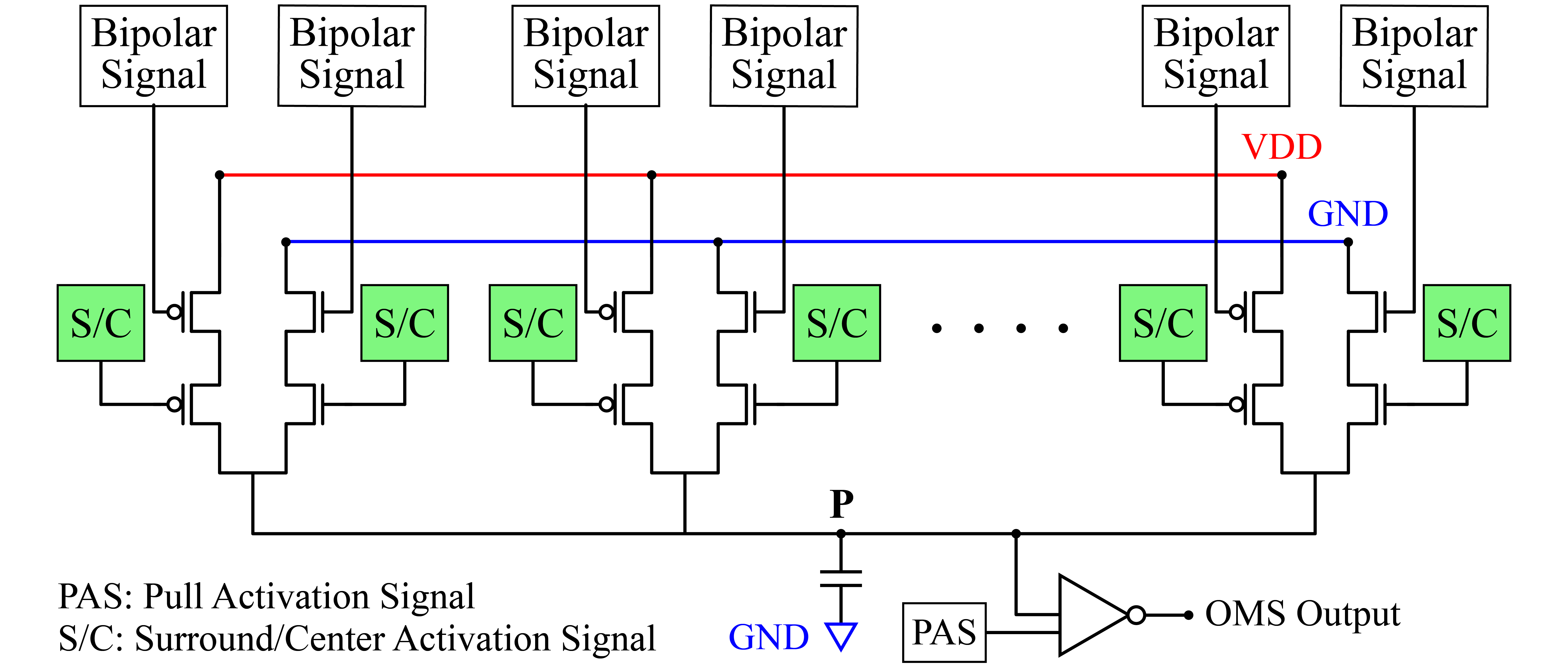}}
    \caption{Schematic of the proposed OMS compute circuit where each pixel can be configured to be a part of the `center' or `surround' region.}
    \label{fig:schematic}
\end{figure}

The proposed run-time reconfigurable circuit is shown in Figure \ref{fig:schematic}. The novel design consists of 2 PMOS and 2 NMOS transistors connected to each DVS pixel. Each pixel can be configured as the part of a center or surround kernel. This allows the required reconfigurabiltiy for the \textbf{kernel size} parameter in the OMS algorithm. To configure a pixel to be a part of the center region, the `Surround or Center (S/C) Activation Signal' will be programmed at `0' (GND) to activate the PMOS and deactivate the NMOS. Consequently, when a Bipolar Signal is generated on a given DVS pixel, the current flows from \si{V_{DD}} and adds charge to the node `P'.  Conversely, applying \si{V_{DD}} to the S/C activates the NMOS transistors while deactivating the PMOS transistors. This configuration enables current to flow from the node `P' to ground, leading to a reduction in overall accumulated charge. By having a control signal, i.e., the S/C, the pixels are no longer constrained to be `center' or `surround' pixels. As a result, the proposed circuit enables the real-time hardware configuration of the pixel's association in the center and surround regions independently, according to the sizes of the center and surround kernels.

\subsection{Reconfigurable Threshold Circuit}
An additional capability of the re-engineered IRIS circuit is a variable inverter. The variable inverter has the functionality of increasing or decreasing the threshold $\tau$ voltage as needed. 

\begin{figure}[h]
\centerline{\includegraphics[width=90mm]{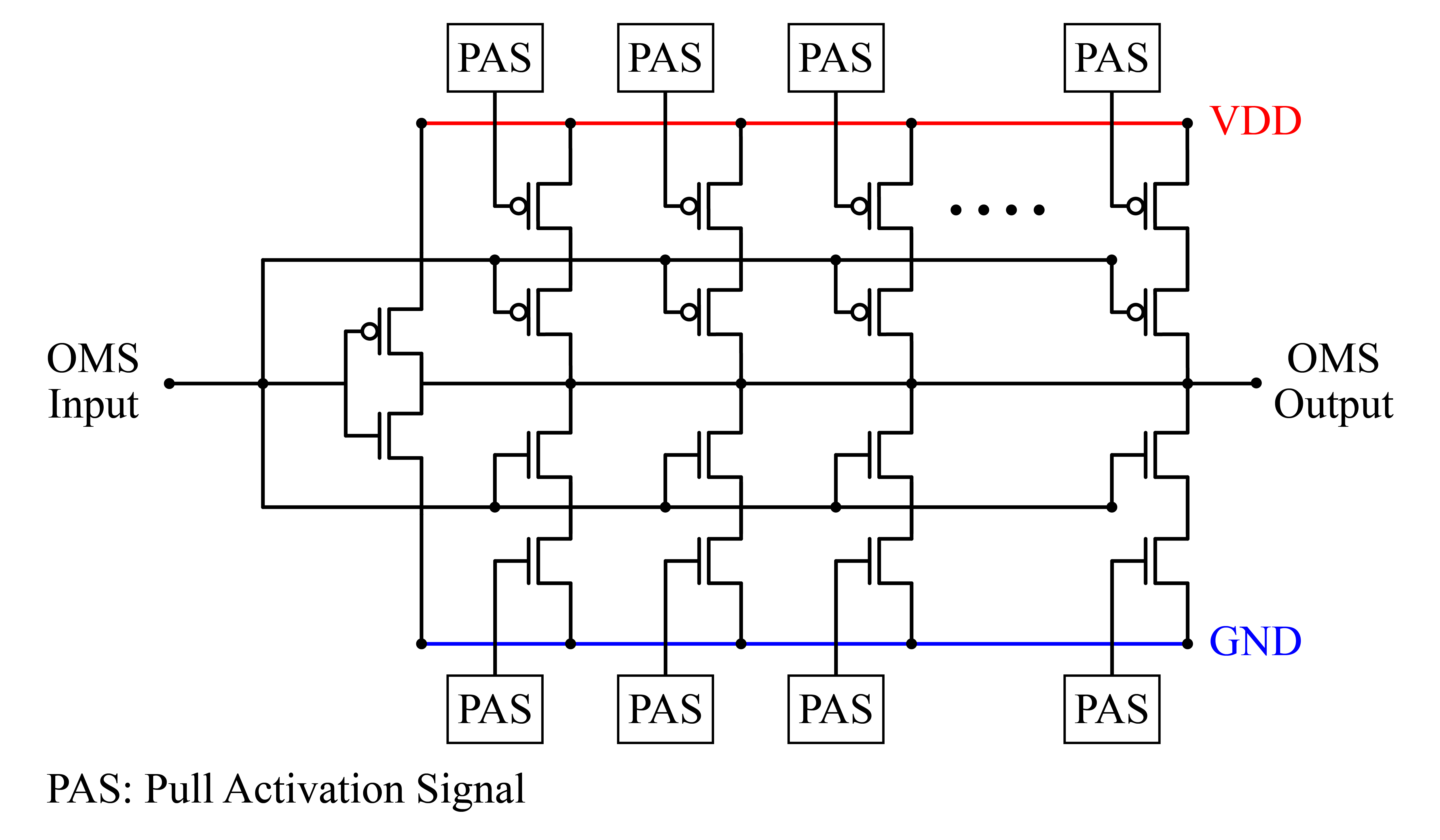}}
\caption{Schematic of the reconfigurable threshold circuit based on programmable inverter's trip point.}
\label{fig:var_inverter}
\end{figure}

\begin{figure*}[!t]
\centerline{\includegraphics[width=0.85\linewidth]{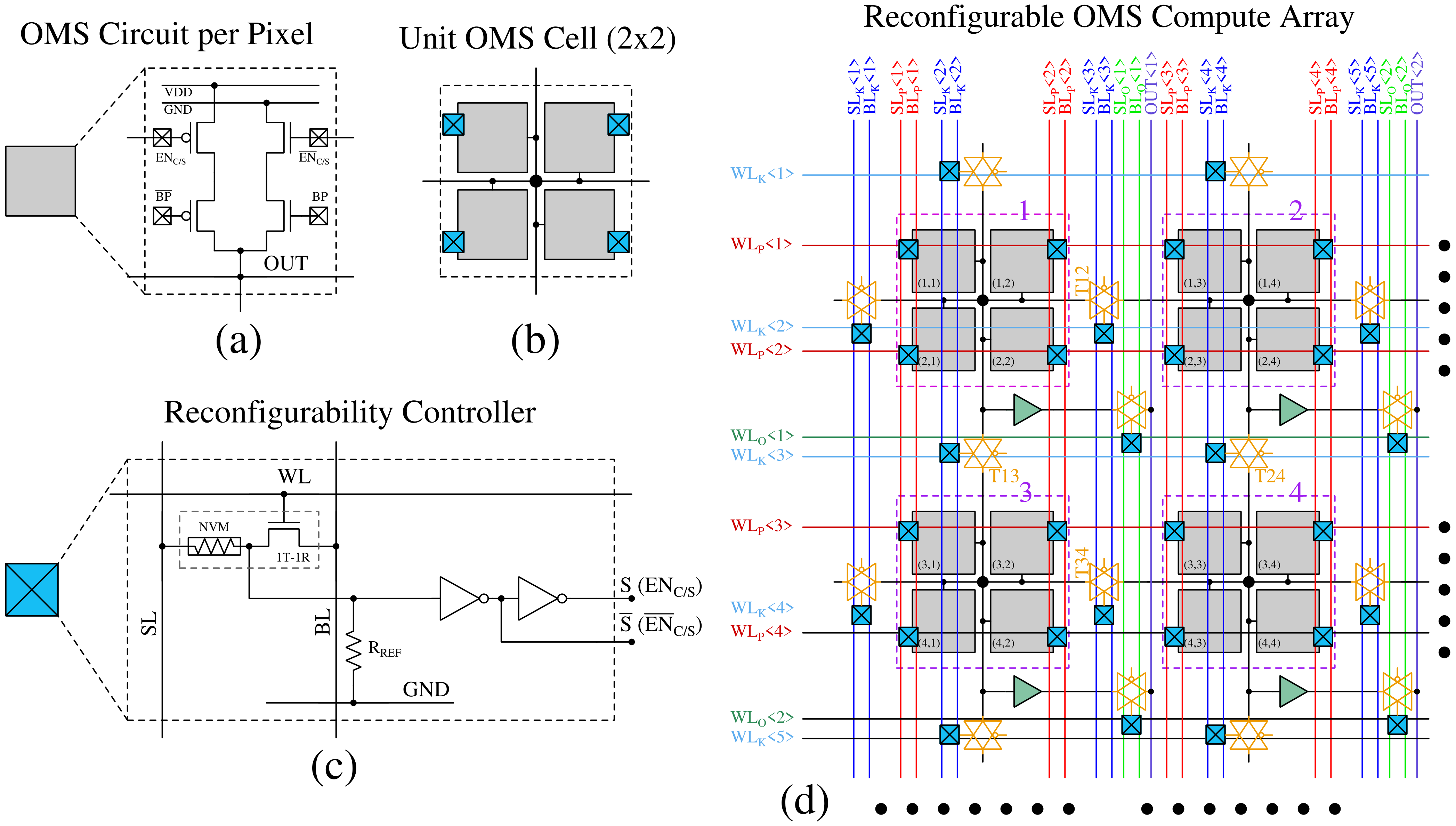}}
\caption{{Reconfigurable OMS Compute Array. (a) OMS compute circuit per pixel, (b) unit OMS compute cell used as the macro block in the 2D array, (c) reconfigurability controller circuit including wordline and bitlines, and (d) reconfigurable OMS compute 2D array.}}
\label{fig:reconfig_array}
\end{figure*}

The inverter circuit shown in Figure \ref{fig:var_inverter} includes a network of pull-up and pull-down transistors. By activating the pull-up transistors utilizing the `PAS' control signals as shown in Figure \ref{fig:var_inverter}, the threshold voltage increases, leading the circuit to produce a logic `0' at higher input voltages. Conversely, activating the pull-down transistors causes the trip voltage to decrease, and therefore, prevents the circuit from producing a logic 0  since a lower input voltage is required. Adding this functionality allows for dynamic changes on the threshold, a key parameter for detecting object motion.
The reconfigurability of the threshold value is crucial for detecting moving objects across various DVS spike densities while achieving high system performance, as outlined in subsection \ref{results_time_window}. During low spike densities (with short time windows), a lower threshold value becomes necessary. Therefore, the flexibility of the threshold range facilitates a hardware-algorithm co-design approach, allowing for short integration times to be utilized.


\subsection{Reconfigurable OMS Compute Array}

Figure \ref{fig:reconfig_array} illustrates the reconfigurable OMS compute array including detailed building blocks, which can program the dimensions of the center and surround (kernel size) region. In Figure \ref{fig:reconfig_array}(a), the proposed OMS compute circuit per pixel is depicted (gray rectangle), where \si{EN_{C/S}} controls the pixel's allocation within a particular center or the surround region. In the case where \si{EN_{C/S}} = GND (VDD), the pixel is considered as a part of the center (surround) region, and connected to the VDD (GND) accordingly. 

`BP' denotes the bipolar signal generated from the image sensor pixel due to the optical intensity change (motion) in the scene. The `OUT' nodes per pixel of the same kernel (center and surround regions) are shorted together and connected to the variable threshold inverter (shown in Figure \ref{fig:var_inverter}). Figure \ref{fig:reconfig_array}(b) shows the unit OMS compute cell (2x2 configurations), arranged in rows and columns within the 2D compute array. The unit cell represents the minimum reconfigurability resolution (in terms of pixel size) in the OMS compute array. 

In addition, the reconfigurability controller circuit (blue rectangle with cross) has been shown in Figure \ref{fig:reconfig_array}(c). First, for the 1T-1R configuration, a non-volatile memory (NVM) device (e.g., MRAM, RRAM, Memristor, etc.) can be employed to store the required setting (center/surround control signals, enable signals for the transmission gates used in the array). The NVM device can store binary data into its resistance state (high or low resistance) and the output control signals S and \si{\overline{S}} can be generated based on the resistance state. For instance, if the NVM is in the high (low) resistance state, S and \si{\overline{S}} becomes VDD (GND) and GND (VDD), respectively. To program the resistance state, the appropriate voltage/current pulse through the BL and SL depending on the NVM device can be supplied, and WL can be utilized to access/program the specific NVM device inside the array. 

Figure \ref{fig:reconfig_array}(d) illustrates the representative diagram of the reconfigurable OMS compute 2D array, including the control lines such as wordlines and bitlines, control switches represented by transmission gates (orange color, e.g., T12, T13, T24, T34, etc.), and OMS compute unit cells labeled as 1, 2, 3, etc. Transmission gates can connect the output node of any two adjacent OMS compute unit cells and can configure the kernel size. Wordlines and bitlines are used to configure the resistance state of the NVM device inside reconfigurability controller and set the appropriate control (\si{EN_{C/S}} to select the center or surround pixel, S to activate the transmission gate) signals. There are three categories of wordlines and bitlines described below:
\begin{itemize}
    \item \textbf{Pixel control:} controls the pixel's association with the center or surround region (indicated by red lines, such as \si{WL_P}$<$1$>$, \si{BL_P}$<$1$>$, \si{SL_P}$<$1$>$, etc.)
    \item \textbf{Kernel size control:} programs the kernel size (center and surround regions in 2x2 resolution) by generating the enable signals of the transmission gates between any two adjacent unit OMS compute cells (shown by blue lines, e.g., \si{WL_K}$<$1$>$, \si{BL_K}$<$1$>$, \si{SL_K}$<$1$>$, etc.).
    \item \textbf{OMS output control:} controls the OMS output propagation (displayed as the green lines, e.g., \si{WL_O}$<$1$>$, \si{BL_O}$<$1$>$, \si{SL_O}$<$1$>$, etc.). 
\end{itemize}

 Each unit OMS compute macro (Figure \ref{fig:reconfig_array}(b)) requires for 5 pairs of bitlines and 5 wordlines to selectively configure the proposed OMS compute array. Hence, effectively each pixel has 2.5 wordlines and 2.5 bitlines, which can easily be supported for the OMS compute circuit without any area overhead as `BP´ generation-circuit is comparatively larger  and hybrid 3D integration provides multiple metal layers to complex routing \cite{yin2022iris}. In addition, the wordlines and bitlines can be driven in parallel. For instance, pixel control wordlines and bitlines can be driven from the west and north side, respectively, and the kernel size control wordlines and bitlines can be driven from the east and south side, respectively) during the programming phase. By programming the reconfigurability controller (blue rectangle with cross) throughout the OMS compute array, we can selectively set the pixels to be a part of the center and surround region as well as configure the center and surround kernel size. Note, the proposed reconfigurable circuits can be integrated in a 3D manner using Cu-Cu hybrid bonding with a backside illuminated image sensor chip as presented in \cite{yin2022iris} for area efficient implementation.

\section{Results}
\subsection{Software Simulations}

For the parameter assessment, we used EV-IMO dataset \cite{mitrokhin2020evimo}. EV-IMO is a widely used dataset for ego-motion compensation captured with a DAVIS 346C DVS camera at 200 Hz. Each video sequence contains binary DVS events with a 346×260 resolution and a 70$^\circ$ field of view. Additionally, motion masks were captured with the \textit{VICON motion capture system}, which generates pixel-wise object masks at 40 Hz. This dataset comprises video sequences captured in controlled environments, with up to three distinct moving objects. Each object is accompanied by a ground truth mask that leverages active pixels to identify the moving objects within each frame. A data sample of the dataset and its ground truth label is shown in Figure~\ref{fig:database}. 

\begin{figure}[h]
    \centering
    \includegraphics[width=0.95\linewidth]{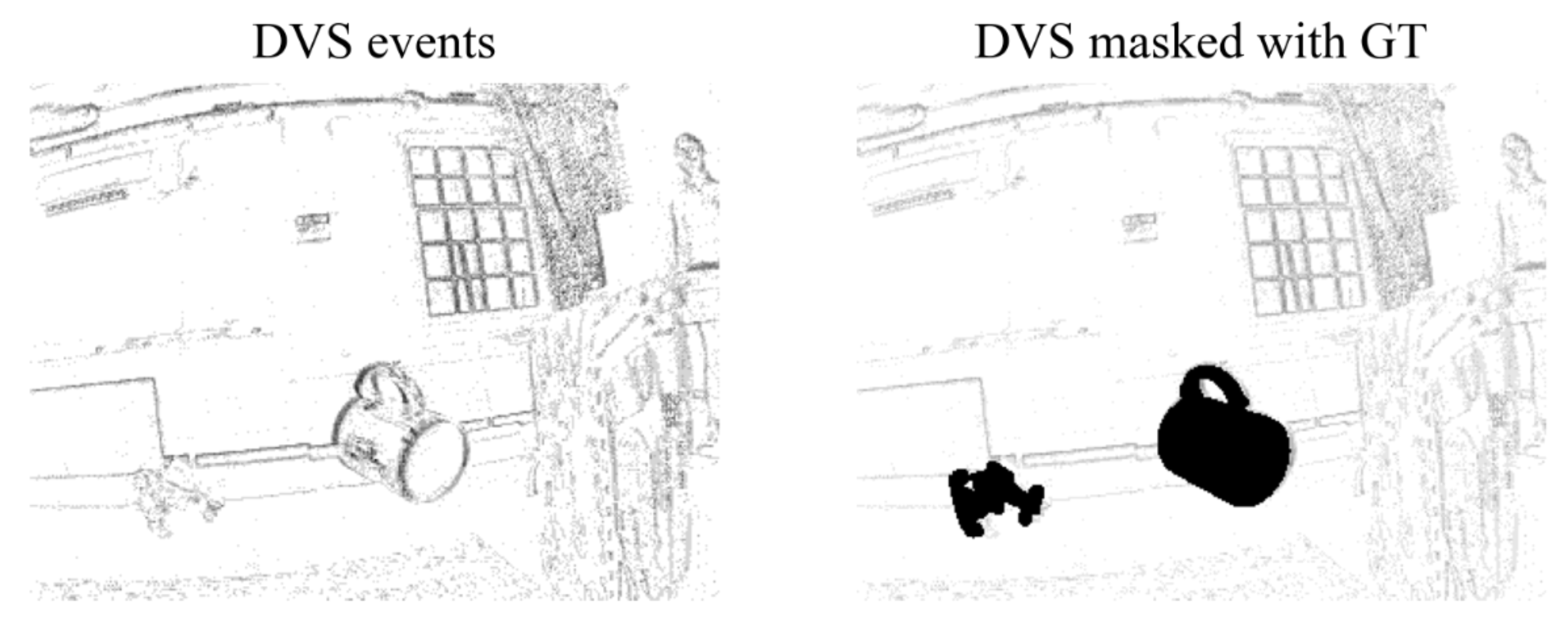}
    \caption{A sample frame and label from EV-IMO \cite{mitrokhin2020evimo} showing a DVS frame and the corresponding ground truth mask.}
    \label{fig:database}
\end{figure}

To evaluate the performance impact of these parameters, we measure the mean \textit{Intersection over Union} (IoU (\%)) between the OMS frames and the corresponding motion mask for a single video sequence. The IoU (\%) measures the degree of overlap between the area of the prediction and the ground truth and is computed as described in \cite{0MMS-IoU}.

\subsubsection{FILTER TYPE}
\begin{figure}[h]
    \centering
    \includegraphics[width=0.95\linewidth]{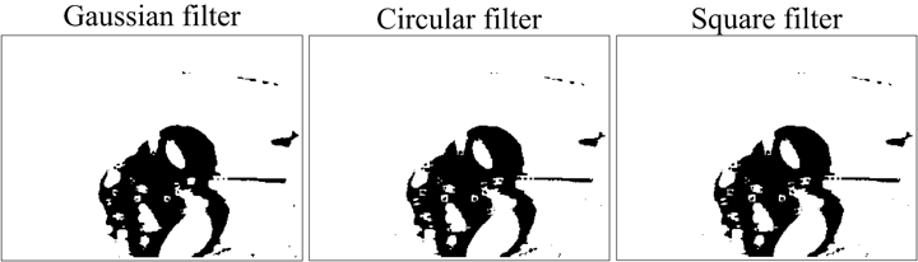}
    \caption{OMS filter type assessment for a sample frame from EV-IMO\cite{mitrokhin2020evimo}. }
    \label{fig:filters}
\end{figure}

As mentioned in the previous section, the original algorithm configuration used a Gaussian filter for the surround and center kernels. However, implementing the sampled Gaussian weights in CMOS technology complicates the design. Hence, we compare the performance of different filter types to select the one that prioritizes the electronic circuitry simplicity and provides a comparable performance in software. Figure \ref{fig:filters} shows the OMS result for a given frame for different kernel types. The visual distinctions observed when computing OMS using various filters are marginal. Nonetheless, from a hardware standpoint, employing a square filter greatly simplifies the circuit design.
\subsubsection{STRIDE}
Table \ref{tab:stride} presents the performance when using fixed kernel sizes with varying strides. The stride refers to the interval by which the filter moves across the input data during the convolution operation. The results show a performance drop of less than 3\%  between convolutions with maximum overlap (stride 1) and zero overlap (stride = kernel size). This suggests that prioritizing a simpler and smaller circuit design (achieved through maximum strides) has  minimal impact on performance. Therefore, we can achieve comparable  performance while favoring a more efficient electronic circuit design.

\begin{table}[h]
\centering
\caption{Stride assessment for a surround kernel shape 8x8 and a center kernel shape 4x4. The mIoU is measured over the wall sequence number 0 from the validation set of EV-IMO \cite{mitrokhin2020evimo}.}
\label{tab:stride}
\begin{tabular}{cc}
\toprule 
 Stride & mIoU(\%)   \\
\midrule
1      & 86.13  \\ 
2      & 83.212 \\
4      & 82.62  \\
6      & 83.98  \\
8      & 83.75 \\
\bottomrule
\end{tabular}
\end{table}

\begin{figure}[h]
    \centering
    \includegraphics[height=3cm]{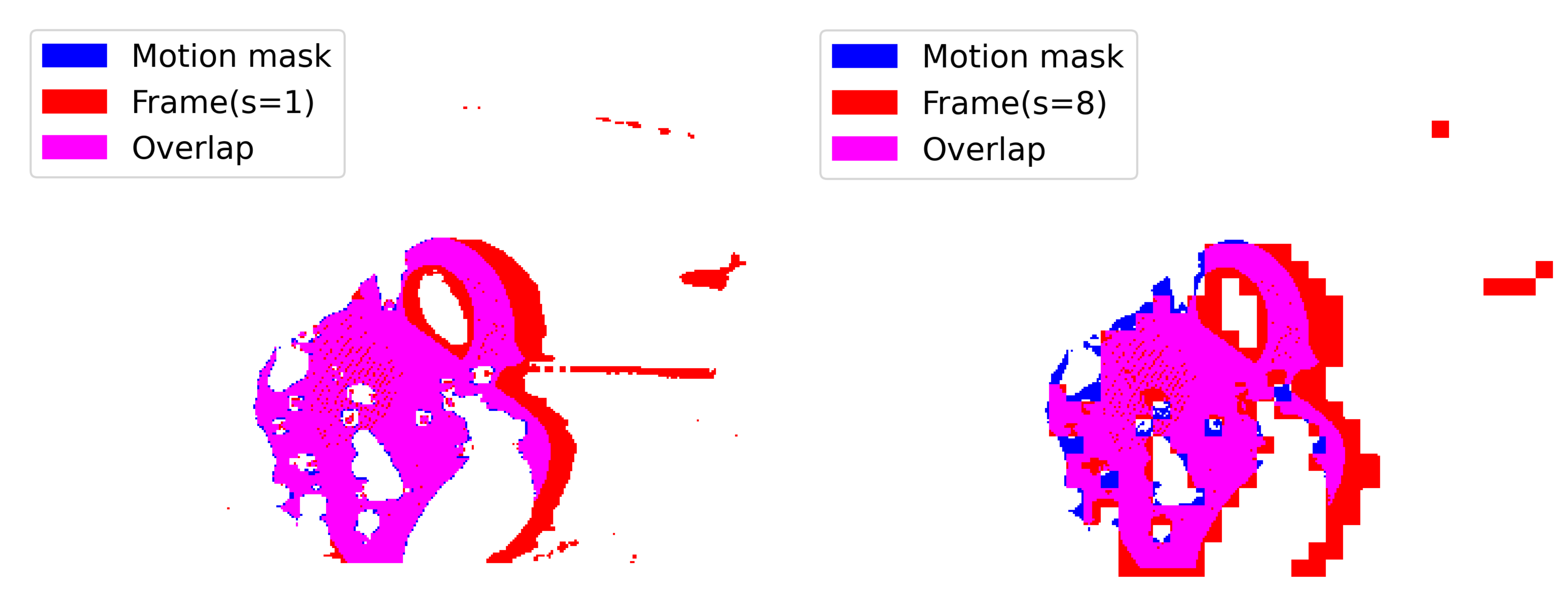}
    \caption{Overlap of the ground truth mask and OMS frames when computing the algorithm with different strides and fixed kernel sizes for a sample frame from EV-IMO\cite{mitrokhin2020evimo}.}
    \label{fig:database}
\end{figure}

\subsubsection{TIME INTERVAL} \label{results_time_window}
\begin{figure}[h]
    \centering
    \includegraphics[width=0.8\linewidth]{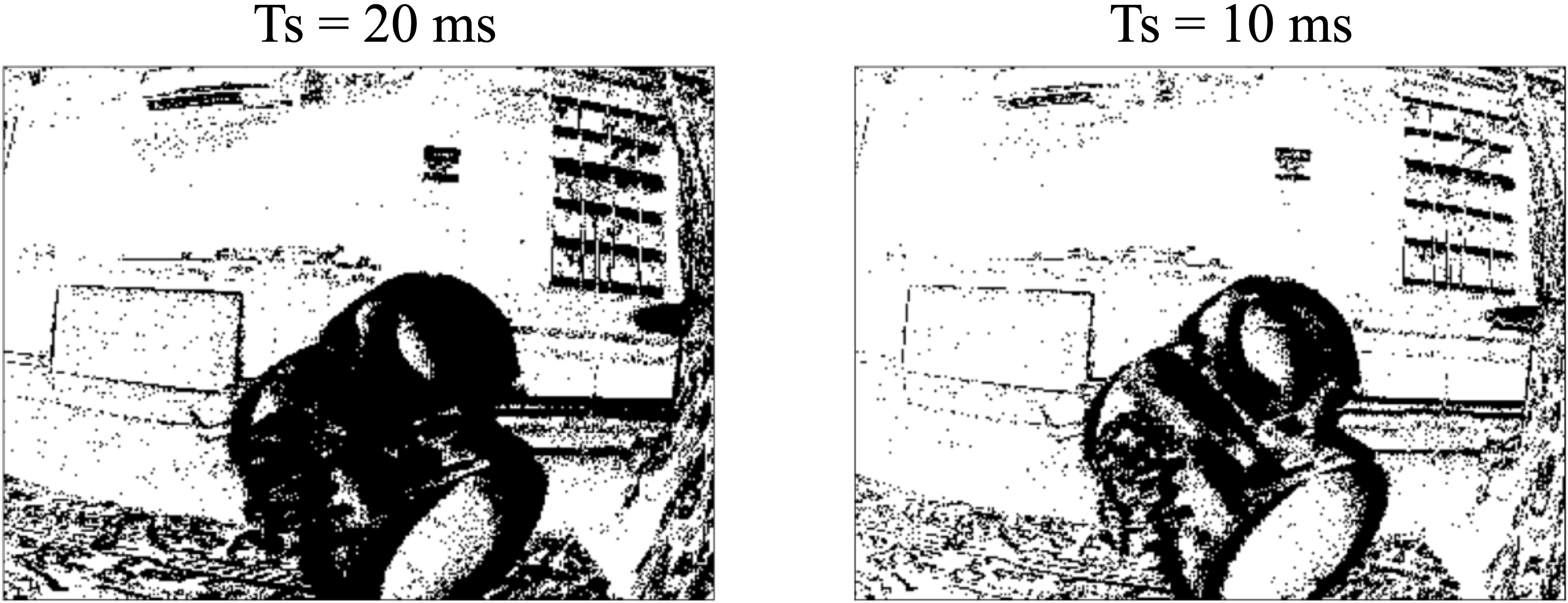}
    \caption{EV-IMO~\cite{mitrokhin2020evimo} frames with time intervals $T_s$ of 20ms and 10ms.}
    \label{fig:time_window}
\end{figure}

\begin{figure}[h]
    \centering
    \includegraphics[width=0.9\linewidth]{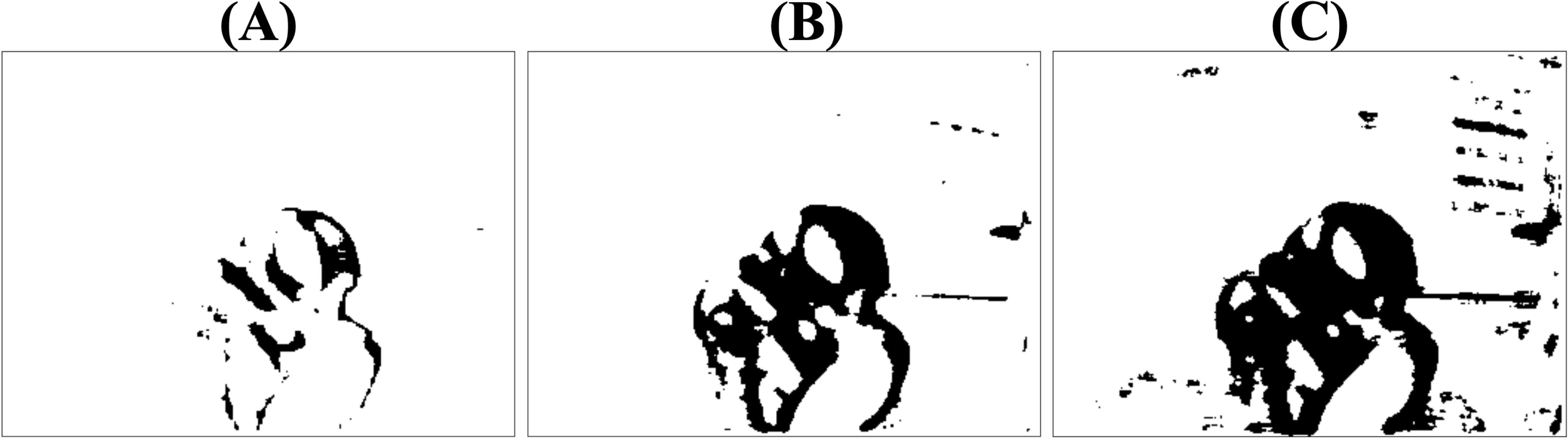}
    \caption{OMS frame using a sample DVS frame from EV-IMO \cite{mitrokhin2020evimo} for a time interval $T_s = 10 ms$ for thresholds (A) $\tau$ = 0.96 (B) $\tau$ = 0.7 and (C) $\tau$ = 0.5}
    \label{fig:time_window2}
\end{figure}

In order to extract the highest amount of spatial information from the DVS events, we accumulate the events for $T_s$ milliseconds around the capture time of the ground truth motion mask. The experiments in previous work \cite{OMS_CHANGE} accumulated the spikes for 20 milliseconds for comparison purposes. 
Due to the intrinsic leakage of CMOS circuits, retaining charges for a long integration time using a simple passive capacitor-based circuit is critical. Hence, a long time interval can generate inaccurate voltage values, consequently degrading the system's IoU. When seeking to decrease $T_s$, the spike density reduces. Figure \ref{fig:time_window}  shows the accumulated DVS frames for time intervals of 20 and 10 milliseconds, i.e. $T_s = 20 ms$ and $T_s = 10 ms$. By condensing the event accumulation period, spatial information achieves greater accuracy, as the relative positions of moving objects exhibit less variation between the initial and final time samples, at the cost of reducing the spike rate.
Computing the OMS algorithm for DVS frames using $T_s = 10 ms$ requires for additional tuning since the spike density decreases. Therefore, it is required to reduce the threshold to maintain the moving object's structure at the output, see Figure \ref{fig:time_window2}. These results show that computing the OMS with a lower integration time is feasible if the threshold is adjusted in consequence, reaffirming the need of a re-configurable design. 

As mentioned, the current hardware design allows reconfiguring the kernel size. Thus, we evaluate the performance of the OMS algorithm for different kernel shapes depending on the relative size of the objects. Relative object sizes are subject to the real object size and to the camera perspective, appearing relatively small if distant from the camera, or large if nearby. When using large kernel sizes, the corresponding cell cluster likely covers a whole moving object if the said object occupies a small portion of the frame's view.
For closer or larger objects, a large kernel can encompass a wider area, capturing a more detailed picture of the object itself. Additionally, it can capture and effectively suppress background or surrounding movement because of its broader view of the scene.
In our experiments, we consider a \textbf{relatively small} object if its area constitutes less than 20\% and \textbf{relatively large }if opposite. 

Results in Table \ref{tab:objects} illustrate the advantages of reconfiguring the sizes of the surround and center regions depending on the relative size of the object, i.e. the size of the object or the relative distance to the camera. This hardware-algorithm adaptability allows for a scalable, robust, and reliable system for object motion detection based on retinal circuits with added benefits of segregating object motion based on object sizes.


\begin{table}[h]
\centering
\caption{Comparison of performance using different kernel sizes for scenes with relatively large and small objects. Kernel sizes are expressed as \textit{center shape }and \textit{center shape} }
\label{tab:objects}
\begin{tabular}{lll}
\toprule
kernel size   &Large objects & Small objects\\
              & $\tau$ = .9            & $\tau$ = .9          \\
              \midrule
3x3 and 6x6   & 75.76      &  85.84          \\
4x4 and 8x8   & 78.59         & 84.70    \\
5x5 and 10x10 & 85.19       &   81.56     \\ 
12x12  and 6x6 & 88.53  & 78.74 \\
\bottomrule
\end{tabular}
\end{table}
\subsection{Hardware Simulations}

Hardware prefers a high stride number (non-overlapping kernel) and a simple square filter to minimize routing complexity and maintain the pixel pitch requirement for high-resolution cameras without the need for additional circuits per pixel. Evaluations of different strides (shown in Table \ref{tab:stride}) and filter types (shown in Figure \ref{fig:filters}) demonstrate minimal impact on system performance  a non-overlapping stride and square filter. Additionally, assessments on time accumulation selection suggest that a short interval can be employed to align with hardware preferences by adjusting the threshold value of the OMS computation. Our proposed programmable threshold circuit can dynamically reconfigure to accommodate software requirements in real-time. In the following subsections, we will discuss the simulations results of our proposed reconfigurable OMS and threshold circuit. 

\begin{figure}[!b]
    \vspace{-0.2in}
    \centerline{\includegraphics[width=1\linewidth]{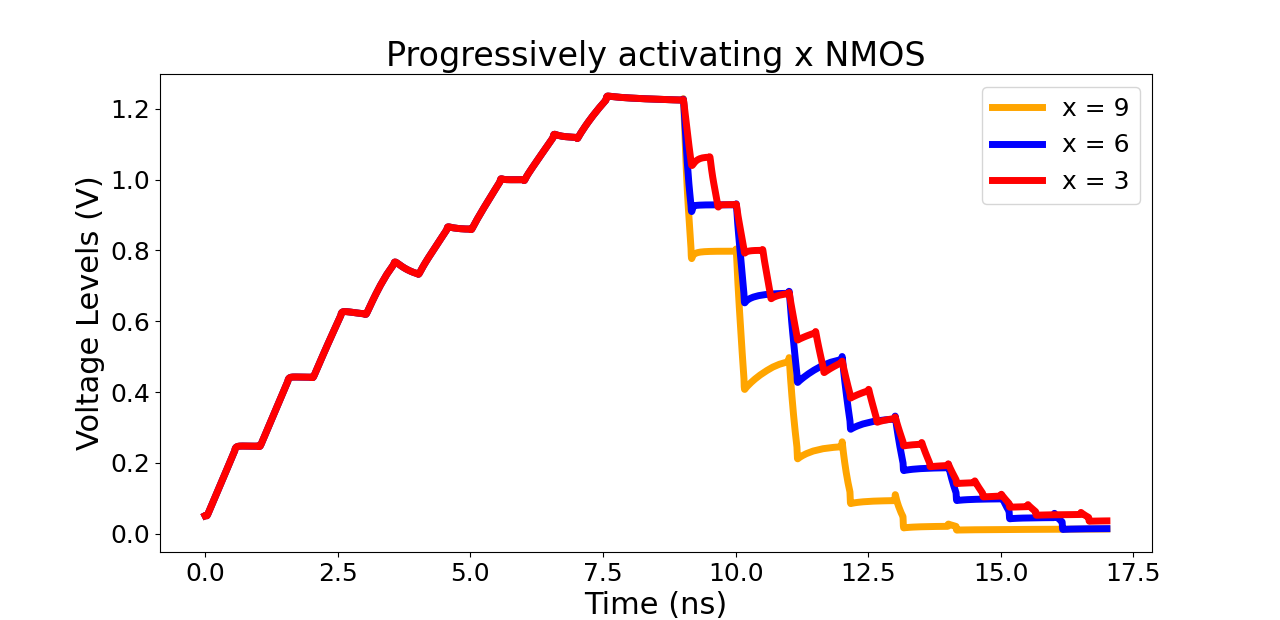}}
    \caption{Voltage levels as different numbers of NMOS and PMOS transistors are being activated.}
    \label{oms_sim}
\end{figure}

\subsubsection{RECONFIGURABLE OMS COMPUTATION CIRCUIT}

We have performed simulations of our proposed circuit in Cadence Virtuoso using 180nm technology node
to validate our hardware design. Figure \ref{oms_sim} shows that the voltage levels on node `P' (as shown in the Figure \ref{fig:schematic}) charges and discharges depending on the bipolar activations of the center and surround regions. For the simulation, 16 pixels were designated as `center' regions and 48 pixels as `surround' regions. Initially, the PMOS transistors were sequentially activated prior to the activation of the NMOS transistors. After that, the NMOS transistors in the surround region are activated sequentially at different numbers, exhibiting different discharging rate (the higher the number of BP activations in the surround pixel, the faster the discharge). Simulations exhibit an average energy per pixel of 9.21 fJ for the three configurations of different number of NMOS transistors being activated, which is a near-negligible overhead compared to pixel energy consumption in typical DVS cameras \cite{kubendran2021256x256}.


\subsubsection{RECONFIGURABLE THRESHOLD CIRCUIT}
The inverter in our circuit is equipped with 10 PMOS transistors and 10 NMOS transistors to provide the pull-up and pull-down networks. We simulated the reconfigurable threshold circuit based on variable trip-points of the modified inverter to observe how the threshold voltage changes with respect to the number of PMOS and NMOS transistors change.

\begin{figure}[!t]
    \centerline{\includegraphics[width=0.75\linewidth]{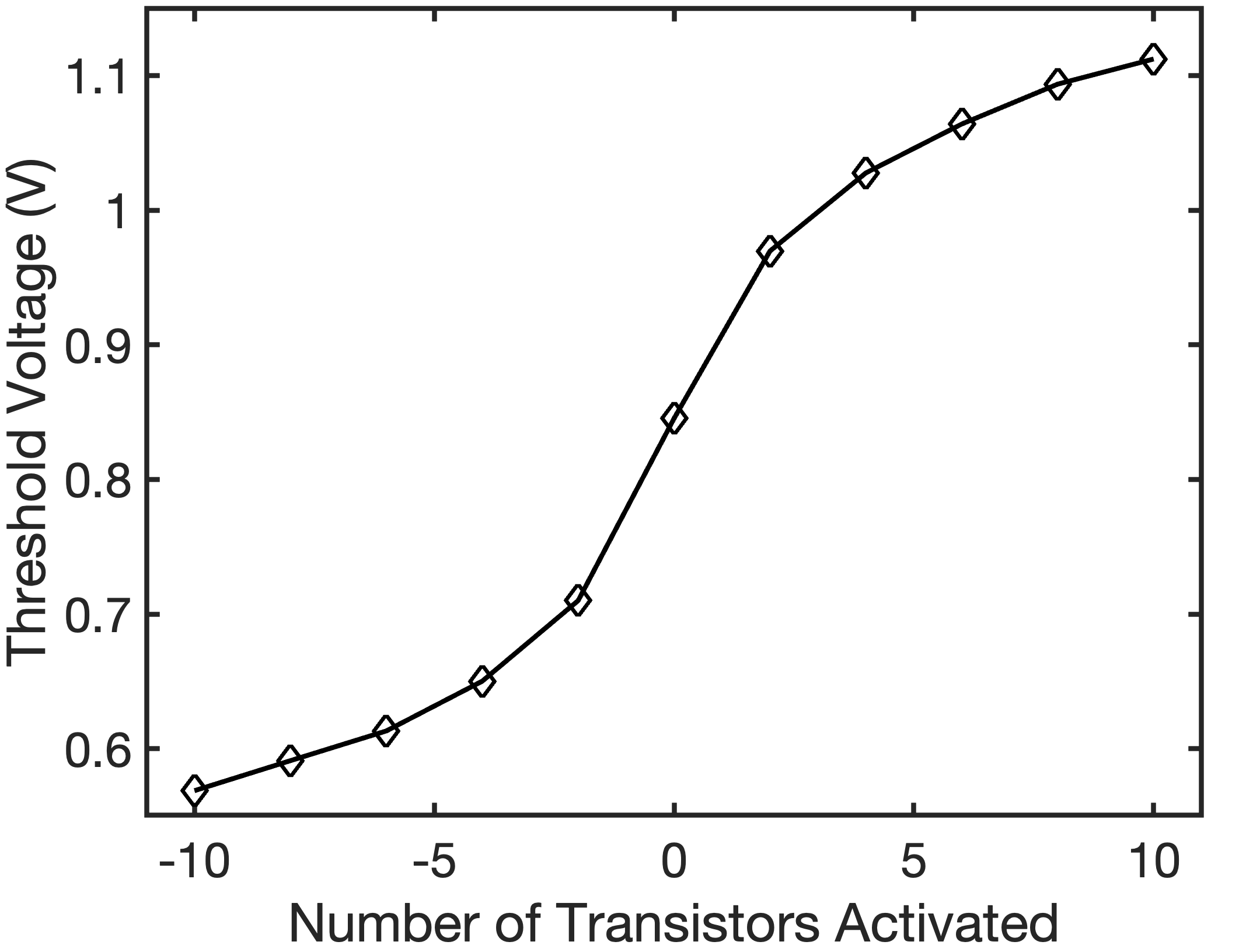}}
    \caption{Variable threshold voltage of our reconfigurable Inverter circuit. Positive (negative) numbers represent the number of PMOS (NMOS) transistors are being activated.}
    \label{inv_sim}
\end{figure}

The threshold voltage of the inverter, as depicted in Figure \ref{inv_sim}, demonstrates reconfigurability ranging from 0.6 V to 1.1 V. Based on the voltage observed at node `P' in the Figure \ref{fig:schematic}, which is dependent upon the center and surround activations, our adaptable threshold inverter can be varied from around 48\% to 90\% of the total voltage range. To support the skewed threshold, we utilize large PMOS transistors to strengthen the pull-up path compared to the pull-down path, shifting the inverter's trip-point closer to \si{V_{DD}} in our reconfigurable threshold circuit. Notably, our algorithm utilizes threshold values within the range of 0.6 to 0.9. Therefore, our achievable hardware threshold range seamlessly aligns with the presented algorithmic requirements.

\section{Discussion \& Conclusion}
This work signifies the convergence of advances made in three distinct fields - retinal neuroscience: by providing detailed insights into biological retinal circuits and their feature extraction behavior; semiconductor 3D integration: allowing 3D integration of multiple stacked chips and using hybrid bonding techniques that have been leveraged to propose functional OPL stacked on top of IPL layers in CMOS circuits; algorithms: that have developed neuroscience inspired algorithmic models which can be used in conjunction with computer vision datasets to study end application accuracy and performance. Interestingly, this neuroscience-hardware-algorithm framework paves the pathway to potentially re-engineer known biological circuits for added functionality, like the retinal OMS circuit studied in this work, to optimize for a given application while keeping in consideration computational and hardware constraints. Similar, cross-stack re-engineering can be performed on several other key biological circuits. Thus, the presented work lays the foundation for investigating biological circuits with an eye on potential application-driven re-engineering.

Further, we would like to emphasize that several variants of the proposed hardware are possible. For example, accumulation time can be improved using known leakage reduction techniques \cite{leakage_cmos} or innovative use of non-volatile memory technologies \cite{leakage_nvm}. In the current work, accumulation time is limited by the frame rate of the DVS dataset. To ensure the proposed circuit does not ever encounter static current flow (leading to static power dissipation) between the center (PMOS transistors) and surround (NMOS transistors), a global center and surround enable signal can be routed to each pixel ensuring the NMOSes and PMOSes are activated one after another and not both at the same time. Overlapping center and surround regions can be implemented using interleaved pixels as proposed in \cite{yin2022iris}.

In conclusion, we present a holistic analysis of algorithmic parameters for a retina-inspired object motion segmentation task and their relation to hardware constraints. The results of our analysis guide CMOS retina-inspired circuits implementing retinal OMS functionality inside camera pixels while allowing run-time reconfigurability for key parameters. Our proposed solution enables the ability to segregate object motion based on the relative object size, opening up new avenues for future work in hardware-algorithm re-engineering of other retinal feature extraction circuits.

\section{Acknowledgments}
The research was funded in part by National Science Foundation
through awards CCF2319617 and CCF2319619.
\bibliographystyle{IEEEtran}
\bibliography{IEEEabrv, sample-base}

\end{document}